\documentclass[twoside]{article}

\usepackage[accepted]{aistats2015}

\usepackage{url}
\urldef{\mailsa}\path|remi@lebret.ch|
\urldef{\mailsb}\path|ronan@collobert.com|

\usepackage{hyperref}
\usepackage{amsfonts,amstext,amsmath,amssymb}

\usepackage{booktabs}
\usepackage{array,multirow,textcomp}

\usepackage{graphicx}
\usepackage{caption}
\usepackage{subcaption}

\usepackage[round,comma,authoryear]{natbib}

\begin{document}

%

%

\twocolumn[

\aistatstitle{Rehabilitation of Count-based Models for Word Vector Representations}

\aistatsauthor{ R\'emi Lebret$^{1,2}$ \and Ronan Collobert$^{1}$ }

\aistatsaddress{ $^{1}$Idiap Research Institute, Martigny, Switzerland\\
$^{2}$Ecole Polytechnique F\'ed\'erale de Lausanne (EPFL), Lausanne, Switzerland\\
\mailsa,
\mailsb} ]

\begin{abstract} 
Recent works on word representations mostly rely on predictive models.
Distributed word representations (aka word embeddings) are trained to optimally predict the contexts in which the corresponding words tend to appear. Such models have succeeded in capturing word similarities as well as semantic and syntactic regularities.
Instead, we aim at reviving interest in a model based on counts.
We present a systematic study of the use of the Hellinger distance to extract semantic representations from the word co-occurrence statistics of large text corpora.
We show that this distance gives good performance on word similarity and analogy tasks, with a proper type and size of context, and a dimensionality reduction based on a stochastic low-rank approximation.
Besides being both simple and intuitive, this method also provides an encoding function which can be used to infer unseen words or phrases. This becomes a clear advantage compared to predictive models which must train these new words.
\end{abstract}

\section{INTRODUCTION}

Linguists assumed long ago that words occurring in similar contexts tend to have similar meanings~\citep{harris54,firth57synopsis}.
Using the word co-occurrence statistics is thus a natural choice to embed similar words into a common vector space~\citep{Turney2010,pennington2014glove}.
Common approaches calculate the frequencies, apply some transformations (tf-idf, PPMI), reduce the dimensionality and calculate the similarities~\citep{Lowe2001}.
Considering a fixed-sized word dictionary $\mathcal{D}$ and a set of words $\mathcal{W}$ to embed, the co-occurrence matrix $C$ is of size $|\mathcal{W}| \times |\mathcal{D}|$. 
$C$ is then dictionary size-dependent.
One can apply a dimensionality reduction operation to $C$ leading to $\bar{C}\in \mathbb{R}^{|\mathcal{W}| \times d}$, where $d \ll |\mathcal{D}|$.
Dimensionality reduction techniques such as Singular Value Decomposition (SVD) are widely used (e.g. LSA~\citep{Landauer1997}, ICA~\citep{Vayrynen04STeP2004}).
In \citet{Bullinaria07extractingsemantic,Bullinaria12extractingsemantic}, the authors provide a full range of factors to use for properly extracting semantic representations from the word co-occurrence statistics of large text corpora.
While word co-occurrence statistics are discrete distributions, an information theory measure such as the Hellinger distance seems to be more appropriate than the Euclidean distance over a discrete distribution space.
In this respect, \citet{Lebret14} propose to perform a principal component analysis (PCA) of the word co-occurrence probability matrix to represent words in a lower dimensional space, while minimizing the reconstruction error according to the Hellinger distance.
In practice, they just apply a square-root transformation to the co-occurrence probability matrix, and then perform the PCA of this new matrix.
They compare the resulting word representations with some well-known representations on named entity recognition and movie review tasks and show that they can reach similar or even better performance.

This paper proposes an extension of the work of \citet{Lebret14} by investigating the impact of different factors.
\citet{Mikolov2013} show that a subsampling approach to imbalance between the rare and frequent words improves the performance.
Recent approaches for word representation have also shown that large windows of context are helpful to capture semantic information~\citep{Mikolov2013,pennington2014glove}.
While, in \citet{Lebret14}, only the 10,000 most frequent words from the dictionary $\mathcal{W}$ are considered as context dictionary $\mathcal{D}$, we investigate various types of context dictionaries, with only frequent words or rare words, or a combination of both.  
In this previous work, the co-occurrence counts to build $C$ are based on a single context word occurring just after the word of interest.
In this paper, we analyse various sizes of context, with both symmetric and asymmetric windows. 
For deriving low-dimensional vector representations from the word co-occurrence matrix $C$, PCA can be done by eigenvalue decomposition of the covariance matrix $C^{T}C$ or SVD of $C$.
Covariance-based PCA of high-dimensional matrices can lead to round-off errors, and thus fails to properly approximate these high-dimensional matrices in low-rank matrices.
And SVD will generally requires a large amount of memory to factorize such huge matrices.
To overcome these barriers, we propose a dimensionality reduction based on stochastic low-rank approximation and show that it outperforms the covariance-based PCA.

Recently, distributed approaches based on neural network language models have revived the field of learning word embeddings~\citep{collobert:2008,huang:2009,TurianRaBe2010,Mnih2013,MikolovICLR2013}.
Such approaches are trained to optimally predict the contexts in which words from $\mathcal{W}$ tend to appear.
\citet{baroni2014} present a systematic comparison of these predictive models with the models based on co-occurrence counts, which suggests that context-predicting models should be chosen over their count-based counterparts. 
In this paper, we aim at showing that count-based models should not be buried so hastily.
A neural network architecture can be hard to train. 
Finding the right hyperparameters to tune the model is often a challenging task and the training phase is in general computationally expensive.
Counting words over large text corpora is on the contrary simple and fast.
With a proper dimensionality reduction technique, word vector representations in a low-dimensional space can be generated.
Furthermore, it gives an encoding function represented by a matrix which can be used to encode new words or even phrases based on their counts.
This is a major benefit compared to predictive models which will need to train vector representations for them.
Thus, in addition to being simple and fast to compute, count-based models become a simple, fast and intuitive solution for inference.

\section{HELLINGER-BASED WORD VECTOR REPRESENTATIONS}

\subsection{Word Co-Occurrence Probabilities}
``You shall know a word by the company it keeps"~\citep{firth57synopsis}.
Keeping this famous quote in mind, word co-occurrence probabilities are computed by counting the number of times each context word $c \in \mathcal{D}$ (where $\mathcal{D}\subseteq\mathcal{W}$) occurs around a word $w \in \mathcal{W}$:
\begin{equation}
p(c|w)=\frac{p(c,w)}{p(w)}=\frac{n(c,w)}{\sum_{c}{n(c,w)}}\,,
\end{equation}
where $n(c,w)$ is the number of times a context word $c$ occurs in the surrounding of the word $w$.
A multinomial distribution of $|\mathcal{D}|$ classes (words) is thus obtained for each word $w$:
\begin{equation}
P_w=\{p(c_1|w),\ldots,p(c_{|\mathcal{D}|}|w)\}\,.
\end{equation}
By repeating this operation over all words from $\mathcal{W}$, the word co-occurrence matrix $C$ is thus obtained:
\begin{equation}
C =  \begin{pmatrix}
  p(c_1|w_1) & \cdots & p(c_{|\mathcal{D}|}|w_1) \\
  p(c_1|w_2) & \cdots & p(c_{|\mathcal{D}|}|w_2) \\
  \vdots  & \ddots & \vdots  \\
  p(c_1|w_{|\mathcal{W}|}) & \cdots & p(c_{|\mathcal{D}|}|w_{|\mathcal{W}|})
  \end{pmatrix}
  = \begin{pmatrix}
  P_{w_1} \\
  P_{w_2} \\
  \vdots \\
  P_{w_{|\mathcal{W}|}}
  \end{pmatrix}
\end{equation}
The number of context words to consider around each word is variable and can be either symmetric or asymmetric. 
The co-occurrence matrix becomes less sparse when this number is high.
Because we are facing discrete probability distributions, the Hellinger distance seems appropriate to calculate similarities between these word representations.
The square-root transformation is then applied to the probability distributions $P_w$, and the word co-occurrence matrix is now defined as:
\begin{equation}
\tilde{C} =   \begin{pmatrix}
  \sqrt{P_{w_1}} \\
  \sqrt{P_{w_2}} \\
  \vdots \\
  \sqrt{P_{w_{|\mathcal{W}|}}}
  \end{pmatrix}
  = \sqrt{C}\,.
\end{equation}

\subsection{Hellinger Distance}
Similarities between words can be derived by computing a distance between
their corresponding word distributions. Several distances (or metrics) over
discrete distributions exist, such as the Bhattacharyya distance, the
Hellinger distance or Kullback-Leibler divergence. We chose here the
Hellinger distance for its simplicity and symmetry property (as it is a
true distance). Considering two discrete probability distributions $P=(p_1,\ldots,p_k)$
and $Q=(q_1,\ldots,q_k)$, the Hellinger distance is formally defined as:
\begin{equation}
H(P,Q)=\frac{1}{\sqrt{2}}\sqrt{\sum_{i=1}^{k}{(\sqrt{p_i}-\sqrt{q_i})^2}}\,,
\end{equation}
which is directly related to the Euclidean norm of the difference of the square root vectors: 
\begin{equation}
H(P,Q)=\frac{1}{\sqrt{2}}\lVert\sqrt{P}-\sqrt{Q}\rVert_2\,.
\end{equation}
Note that it makes more sense to take the Hellinger distance rather than
the Euclidean distance for comparing discrete distributions, as $P$ and $Q$
are unit vectors according to the Hellinger distance ($\sqrt{P}$ and
$\sqrt{Q}$ are units vector according to the $\ell_2$ norm).

\subsection{Dimensionality Reduction}
\label{dim-reduc}
As discrete distributions are dictionary size-dependent, using
directly the distribution as a word representation is, in general, not really tractable for
large dictionary. 
This is even more true in the case of a large number of context words, distributions becoming less sparse.
We investigate two approaches to embed these representations in a low-dimensional space: (1) a principal component analysis (PCA) of the word co-occurrence matrix $\tilde{C}$, (2) a stochastic low-rank approximation to encode distributions $\sqrt{P_w}$.

\subsubsection{Principal Component Analysis (PCA)}
We perform a principal component analysis
(PCA) of the square root of the word co-occurrence probability matrix to
represent words in a lower dimensional space, while minimizing the
reconstruction error according to the Hellinger distance.
This PCA can be done by eigenvalue decomposition of the covariance matrix $\tilde{C}^T\tilde{C}$. With a limited size of context word dictionary $\mathcal{D}$ (tens of thousands of words), this operation is performed very quickly (See~\citet{Lebret14} paper for details).
With a larger size for $\mathcal{D}$, a truncated singular value decomposition of $\tilde{C}$ might be an alternative, even if it is time-consuming and memory-hungry.

\subsubsection{Stochastic Low-Rank Approximation (SLRA)}
When dealing with large dimensions, the computation of the covariance matrix might accumulate floating-point roundoff errors.
To overcome this issue and to still fit in memory, we propose a stochastic low-rank approximation to represent words in a lower dimensional space.
It takes a distribution $\sqrt{P_w}$ as input, encodes it in a more compact representation, and is trained to reconstruct its own input from that representation:
\begin{equation} \label{eq:word}
|| VU^{T}\sqrt{P_w} - \sqrt{P_w} ||^2\,,
\end{equation}
where $U$ and $V \in \mathbb{R}^{|\mathcal{D}|\times d}$.
$U$ is a low-rank approximation of the co-occurrence matrix $\tilde{C}$ which maps distributions in a $d$-dimension (with $d \ll |\mathcal{D}|$), and $V$ is the reconstruction matrix.
$U^{T}\sqrt{P_w}$ is a distributed representation that captures the main factors of variation in the data as the Hellinger PCA does.
$U$ and $V$ are trained by backpropagation using stochastic gradient descent.

\section{EXPERIMENTS}

\begin{figure*}[ht]
\centering
\begin{subfigure}[b]{0.3\textwidth}
        \includegraphics[width=\textwidth]{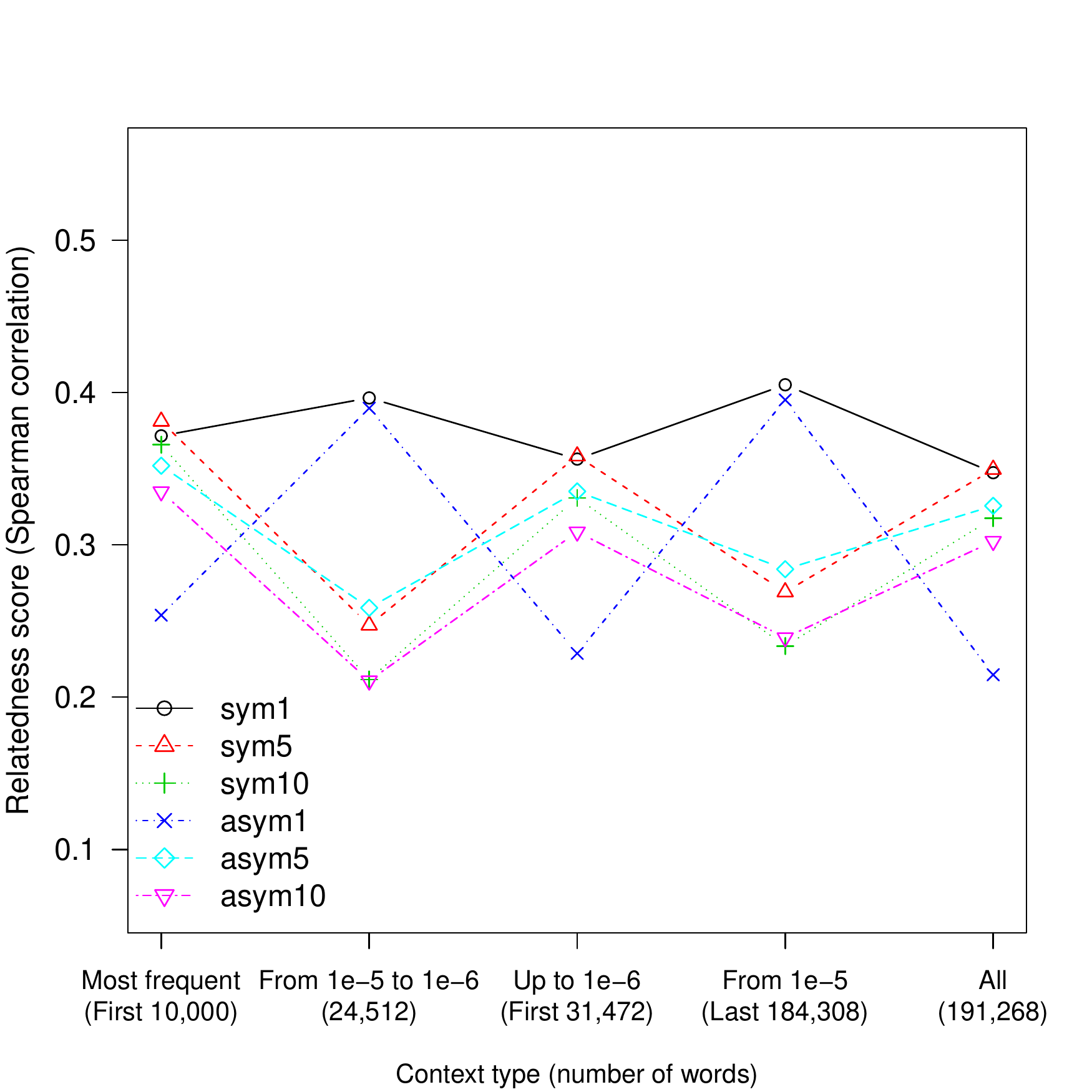}
        \caption{WS-353 dataset}
\end{subfigure}%
~
\begin{subfigure}[b]{0.3\textwidth}
        \includegraphics[width=\textwidth]{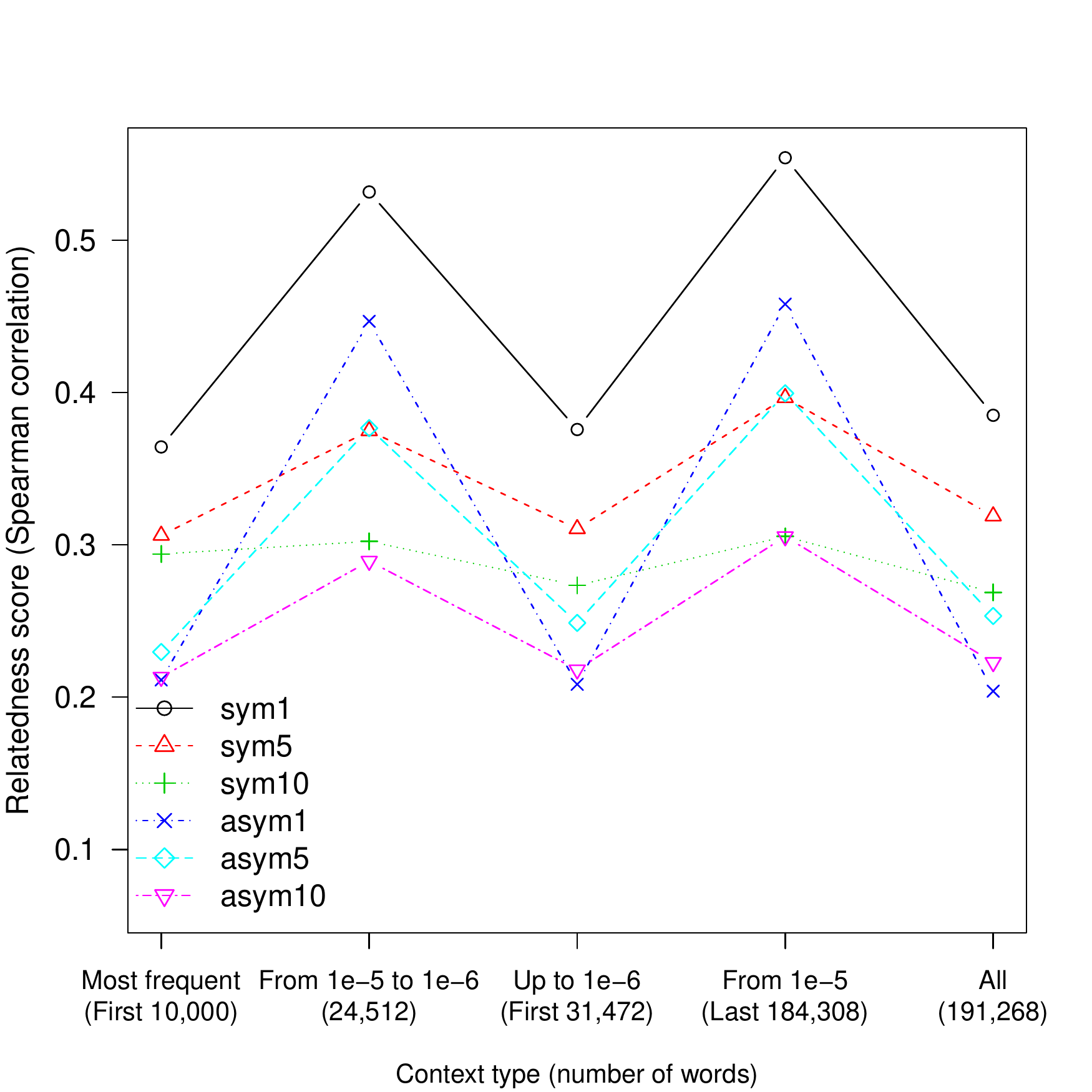}
        \caption{RG-65 dataset}
\end{subfigure}%
~
\begin{subfigure}[b]{0.3\textwidth}
        \includegraphics[width=\textwidth]{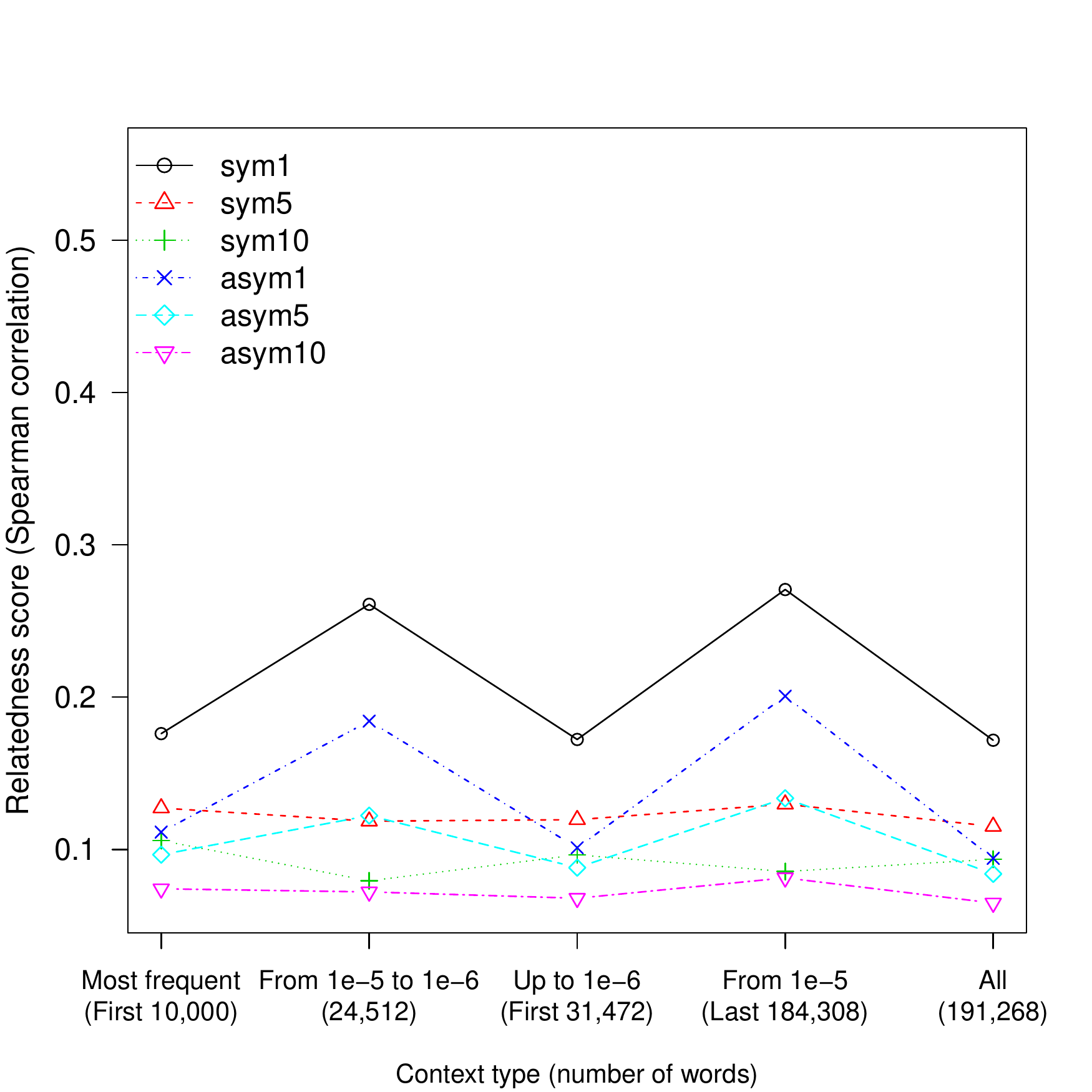}
        \caption{RW dataset}
\end{subfigure}
\begin{subfigure}[b]{0.3\textwidth}
        \includegraphics[width=\textwidth]{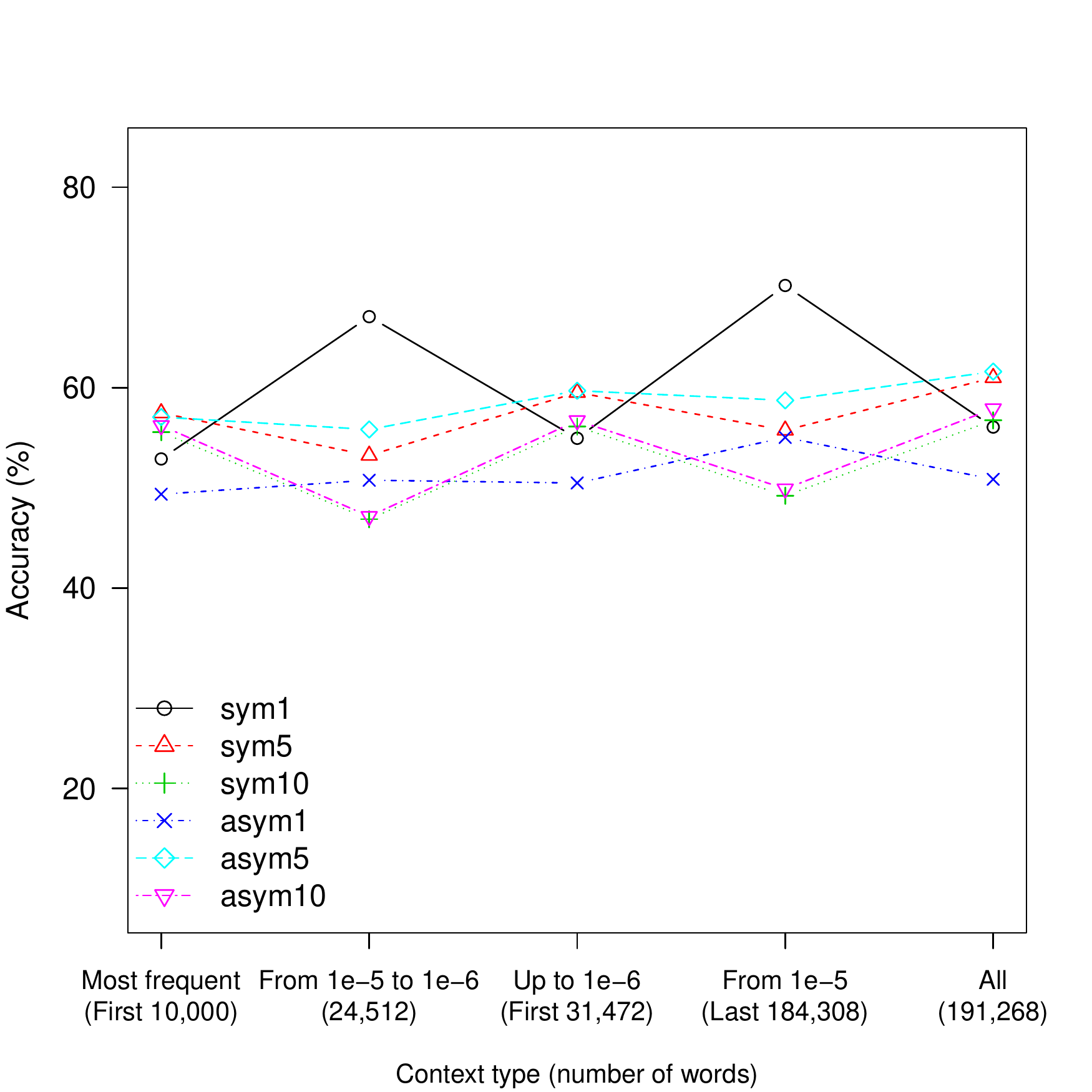}
        \caption{Mikolov's syntactic dataset}
\end{subfigure}%
~
\begin{subfigure}[b]{0.3\textwidth}
        \includegraphics[width=\textwidth]{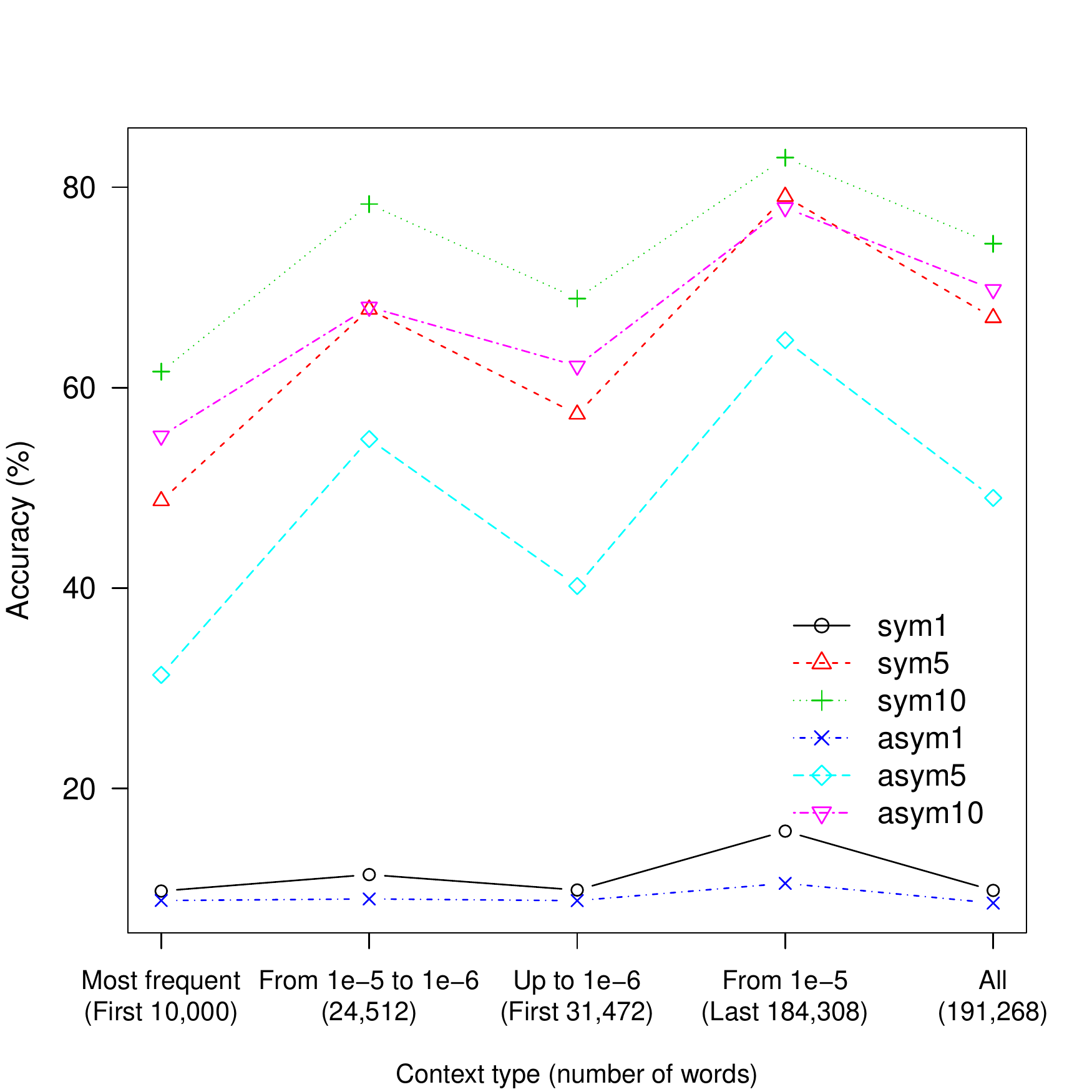}
        \caption{Mikolov's semantic dataset}
\end{subfigure}
\caption{Performance on datasets with different types of context word dictionaries $\mathcal{D}$ (scenarios in the ascending order of their number of words), and different window sizes (in the legend, \emph{sym1} is for symmetric window of 1 context word, \emph{asym1} is for asymmetric window of 1 context word, etc.). Spearman rank correlation is reported on word similarity tasks. Accuracy is reported on word analogy tasks.}
\label{fig:cxt}
\end{figure*}

\subsection{Building Word Representation over Large Corpora}
\label{wordrep}

Our English corpus is composed of the entire English Wikipedia\footnote{Available at \url{http://download.wikimedia.org}. We took the January 2014 version.} (where all MediaWiki markups have been removed). We consider lower case words to limit the number of words in the dictionary. Additionally, all occurrences of sequences of numbers within a word are replaced with the string ``NUMBER". The resulting text is tokenized using the Stanford tokenizer\footnote{Available at \url{http://nlp.stanford.edu/software/tokenizer.shtml}}. The data set contains about 1.6 billion words.
As dictionary $\mathcal{W}$, we consider all the words within our corpus which appear at least one hundred times. 
This results in a 191,268 words dictionary.
Five scenarios are considered to build the word co-occurrence probabilities with context words $\mathcal{D}$:
(1) Only the 10,000 most frequent words within this dictionary.
(2) All the dictionary.
\citet{Mikolov2013} have shown that better word representations can be obtained by subsampling of the frequent words. We thus define the following scenarios:
(3) Only words whose appearance frequency is less than $10^{-5}$, which is the last 184,308 words in $\mathcal{W}$.
(4) To limit the dictionary size, we consider words whose appearance frequency is less than $10^{-5}$ and greater than $10^{-6}$. This results in 24,512 context words.
(5) Finally, only words whose appearance frequency is greater than $10^{-6}$, which gives 31,472 words.

\subsection{Evaluating Word Representations}
\label{dataset}
\subsubsection{Word analogies}
The word analogy task consists of questions like, ``\emph{a} is to \emph{b} as \emph{c} is to ?''. It was introduced in \citet{MikolovICLR2013} and contains 19,544 such questions, divided into a semantic subset and a syntactic subset. The 8,869 semantic questions are analogies about places, like ``\emph{Bern} is to \emph{Switzerland} as \emph{Paris} is to  ?'', or family relationship, like  ``\emph{uncle} is to \emph{aunt} as \emph{boy} is to  ?''. The 10,675 syntactic questions are grammatical analogies, involving plural and adjectives forms, superlatives, verb tenses, etc. To correctly answer the question, the model should uniquely identify the missing term, with only an exact correspondence counted as a correct match.

\subsubsection{Word Similarities}
We also evaluate our model on a variety of word similarity tasks. These include the WordSimilarity-353 Test Collection (WS-353)~\citep{Finkelstein2002}, the Rubenstein and Goodenough dataset (RG-65)~\citep{Rubenstein1965}, and the Stanford Rare Word (RW)~\citep{Luong2013}. They all contain sets of English word pairs along with human-assigned similarity judgements. WS-353 and RG-65 datasets contain 353 and 65 word pairs respectively. Those are relatively common word pairs, like \emph{computer:internet} or \emph{football:tennis}. The RW dataset differs from these two datasets, since it contains 2,034 pairs where one of the word is rare or morphologically complex, such as \emph{brigadier:general} or \emph{cognizance:knowing}.

\subsection{Analysis of the Context}
\label{ana-context}

As regards the context, two main parameters are involved: 
(1) The context window size to consider, \emph{i.e.} the number of context words $c$ to count for a given word $w$. We can either count only context words that occurs after $w$ (asymmetric context window), or we can count words surrounding $w$ (symmetric context window). 
(2) The type of context to use, \emph{i.e.} which words are to be chosen for defining the context dictionary $\mathcal{D}$. Do we need all the words, the most frequent ones or, on the contrary, the rare ones?
Figure \ref{fig:cxt} presents the performance obtained on the benchmark datasets for all the five scenarios described in Section \ref{wordrep} with different sizes of context.
No dimensionality reduction has been applied in this analysis. 
Similarities between words are calculated with the Hellinger distance between the word probability distributions. 
For the word analogy task, we used the objective function \textsc{3CosMul} defined by \citet{Levy2014}, as we are dealing with explicit word representations in this case.

\begin{table}[h]
\begin{center}
\begin{tabular}{@{}lcc@{}}\hline\toprule
& \multicolumn{2}{c}{\bf WINDOW SIZE} \\
\cmidrule{2-3}
& 1 & 10 \\\bottomrule
 \\
\multirow{2}{2.5cm}{{\bf \sc baikal} (n\textsuperscript{o}37415)} & \sc\small m\"alaren & \sc\small lake  \\
 & \sc\small titicaca & \sc\small siberia \\
 & \sc\small balaton & \sc\small amur  \\
 & \sc\small ladoga & \sc\small basin  \\
 & \sc\small ilmen & \sc\small volga   \\\bottomrule
 \\
\multirow{2}{2.5cm}{{\bf \sc special-need} (n\textsuperscript{o}165996)} & \sc\small at-risk & \sc\small preschool \\
 & \sc\small school-age & \sc\small kindergarten \\
 & \sc\small low-income & \sc\small teachers \\
 & \sc\small hearing-impaired & \sc\small schools \\
 & \sc\small grade-school & \sc\small vocational \\\bottomrule
\end{tabular}
\end{center}
\caption{Two rare words with their rank and their 5 nearest words with respect to the Hellinger distance, for a symmetric window of 1 and 10 context words.}
\label{neigh-table}
\end{table}

\begin{table}[h]
\begin{center}
\begin{tabular}{@{}lcccc@{}}\hline\toprule
{\bf TYPE} & {\bf DIM.}& \multicolumn{3}{c}{\bf SIZE}\\
\cmidrule{3-5}
 & & 1 & 5 & 10 \\\bottomrule
 \\
Most frequent & 10000 & 297 & 1158 & 1618  \\
From $10^{-5}$ to $10^{-6}$ & 24512 & 132 & 674 & 1028 \\
Up to $10^{-6}$ & 31472 & 396 & 1672 & 2408 \\
From $10^{-5}$ & 184308 & 249 & 1305 & 2050 \\
All  & 191268 & 513 & 2304 & 3430 \\\bottomrule
\hline
\end{tabular}
\end{center}
\caption{The average number of context words in the co-occurrence matrix according to the type and the size of context.}
\label{cxt-table}
\end{table}

\subsubsection{Window Size}
Except for semantic analogy questions, best performance are always obtained with symmetric context window of size 1. However, performance dramatically drop with this window size on the latter. 
It seems that a limited window size helps to find syntactic similarities, but a large window is needed to detect the semantic aspects. The best results are thus obtained with a symmetric window of 10 words on the semantic analogy questions task.
This intuition is confirmed by looking at the nearest neighbors of certain rare words with different sizes of context.  
In Table \ref{neigh-table}, we can observe that a window of one context word brings together words that occur in a same syntactic structure, while a window of ten context words will go beyond that and add semantic information. With only one word of context, Lake \emph{Baikal} is therefore neighbor to other lakes, and the word \emph{special-needs} is close to other words composed of two words. 
With ten words of context, the nearest neighbors of \emph{Baikal} are words in direct relation to this location, \emph{i.e.} these words cannot match with other lakes, like Lake \emph{Titicaca}. This also applies for the word \emph{special-needs}, where we find words related to the educational meaning of this word.
This could explain why the symmetric window of one context word gives the best results on the word similarity and syntactic tasks, but performs very poorly on the semantic task. 
Finally, the use of a symmetric window instead of an asymmetric one always improves the performance.

\subsubsection{Type of Context}
First, using all words as context does not imply to reach the best performance.
With the 10,000 most frequent words, performance are fairly similar than with all words.
An in-between situation with words whose appearance frequency is greater than $10^{-6}$ gives also quite similar performance.
Secondly, discarding the most frequent words from the context distributions helps, in general, to increase performance. 
The best performance is indeed obtained with scenarios (3) and (4).
But all rare words are not necessarily essential to achieve good performance, since results with words whose appearance frequency is less than $10^{-5}$ and greater than $10^{-6}$ are not significantly lower. 
These two observations might be explained by the sparsity of the probability distributions. 
Counts in Table \ref{cxt-table} show significant differences in terms of sparsity depending on the type of context. 
Similarities between words seem to be easier to find with sparse distributions.
The average number of context words (\emph{i.e.} features) whose appearance frequency is less than $10^{-5}$ and greater than $10^{-6}$ with a symmetric window of size 1 is extremely low (132). Performance with these parameters are still highly competitive on syntactic tasks. Within this framework, it then becomes a good option for representing words in a low and sparse dimension.

\begin{figure*}[ht]
\centering
\begin{subfigure}[b]{0.5\textwidth}
        \includegraphics[width=\textwidth]{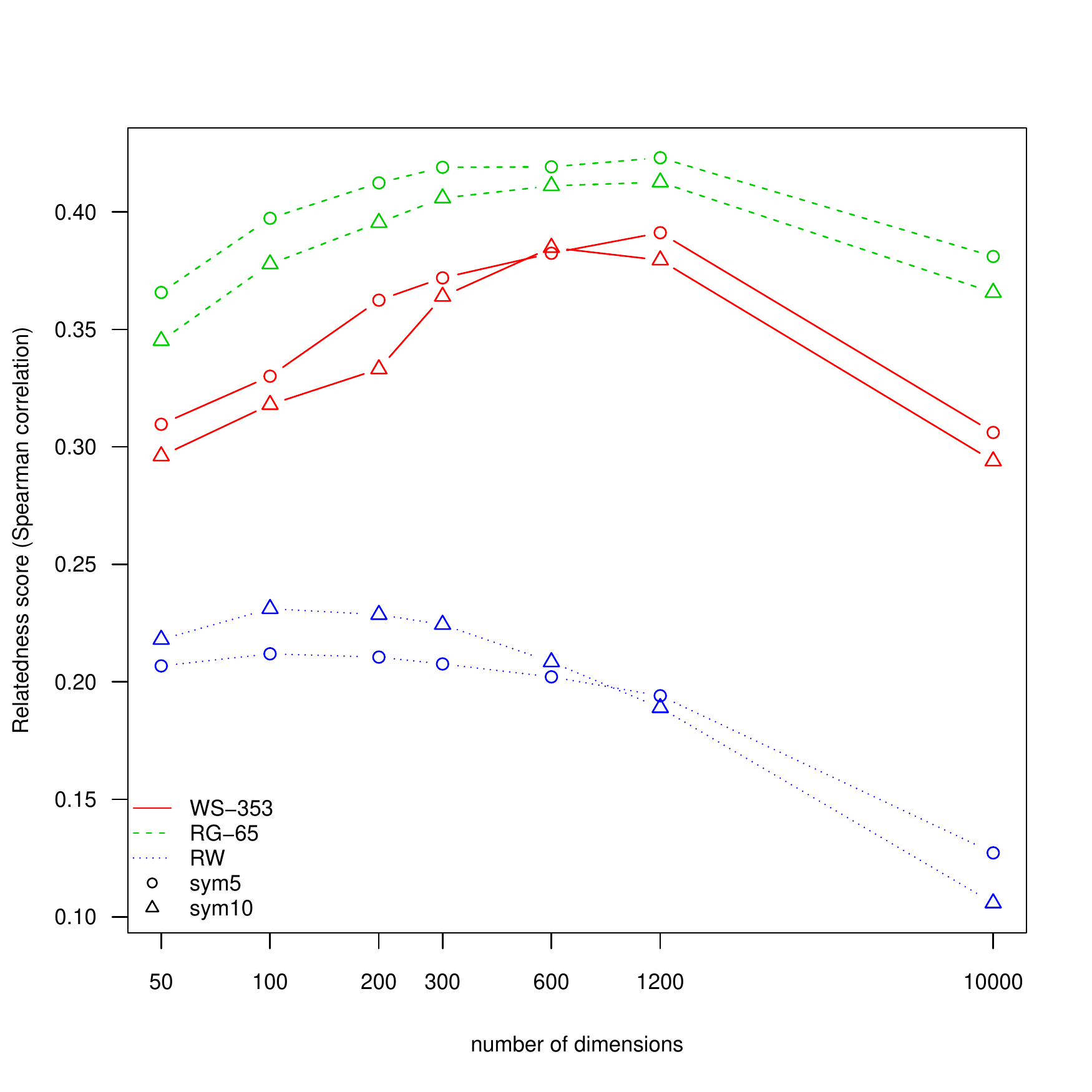}
        \caption{Word similarities datasets}
\end{subfigure}%
~
\begin{subfigure}[b]{0.5\textwidth}
        \includegraphics[width=\textwidth]{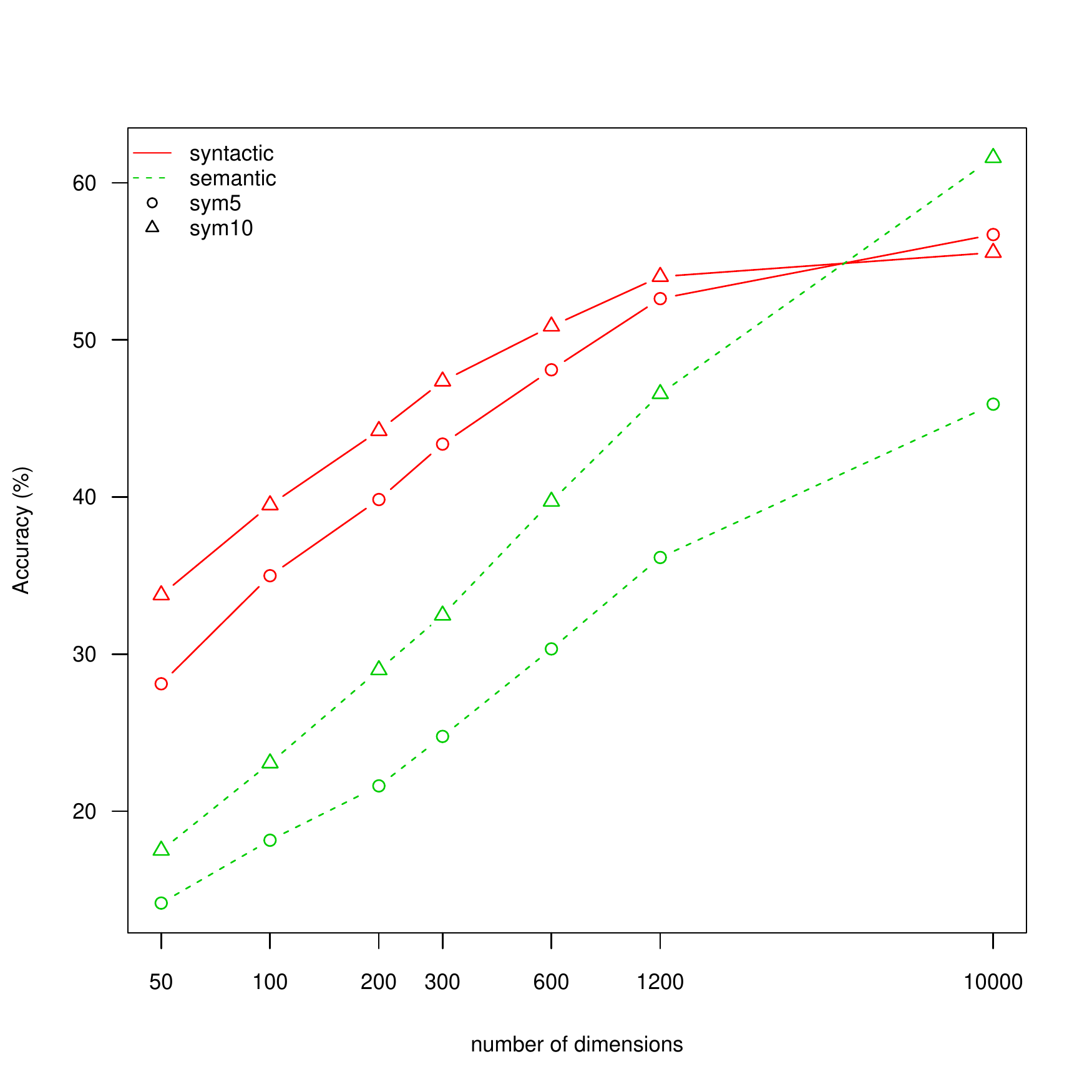}
        \caption{Word analogies datasets}
\end{subfigure}
\caption{Performance on datasets with different dimensions using scenario (1). Dimensionality reduction has been obtained with the Hellinger PCA. Spearman rank correlation is reported on word similarity tasks. Accuracy is reported on word analogy tasks.}
\label{fig:dim}
\end{figure*}

\subsection{Dimensionality Reduction Models}
The analysis of the context reveals that word similarities can even be found with extremely sparse word vector representations. But these representations lack semantic information since they perform poorly on the word analogy task involving semantic questions.
A symmetric window of five or ten context words seems to be the best options to capture both syntactic and semantic information about words.
The average number of context words is much larger within these parameters, which justifies the need of dimensionality reduction.
Furthermore, this analysis show that a large number of context words is not necessary to achieve significant improvements.
Good performance on syntactic and similarity tasks can be reached with the 10,000 most frequent words as context.
Using instead a distribution of a limited number of rare words increases performance on the semantic task while reducing performance on syntactic and similarity tasks.
We then focus on the two scenarios with the fewest number of context words: scenarios (1) and (4) with 10,000 and 24,512 words respectively. 
This reasonable number of context words allows for dimensionality reduction methods to be applied in an efficient manner.

\subsubsection{Number of Dimensions}

When a dimensionality reduction method is applied, a number of dimensions needs to be chosen.
This number has to be large enough to retain the maximum variability. 
It also has to be small enough for the dimensionality reduction to be truly meaningful and effective.  
We thus analyse the impact of the number of dimensions using the Hellinger PCA of the co-occurrence matrix from scenario (1) with a symmetric context of five and ten words.
Figure \ref{fig:dim} reports performance on the benchmark datasets described in Section \ref{dataset} for different numbers of dimensions. 
The ability of the PCA to summarize the information compactly leads to improved results on the word similarity tasks, where performance is better than with no dimensionality reduction.
On the WS-353 and RG-65 datasets, we observe that the gain in performance tends to stabilize between 300 and 1,200 dimensions.
The increase in dimension leads to a small drop after 100 dimensions on the RW dataset.
However, adding more and more dimensions helps to increase performance on word analogy tasks, especially for the semantic one.
We also observe that ten context words instead of five give better results for word analogy tasks, while the opposite is observed for word similarity tasks. This confirms the results observed in Section \ref{ana-context}.

\begin{table*}[ht]
\begin{center}
\begin{tabular}{@{}lccccccccccccc@{}}\hline\toprule
 & \multicolumn{6}{c}{\bf SLRA} & \phantom{a} & \multicolumn{6}{c}{\bf HPCA}\\
 \cmidrule{2-7}  \cmidrule{9-14}
Dimension & \multicolumn{2}{c}{100} & \multicolumn{2}{c}{200} & \multicolumn{2}{c}{300} & & \multicolumn{2}{c}{100} & \multicolumn{2}{c}{200} & \multicolumn{2}{c}{300} \\
 \cmidrule{2-3}  \cmidrule{4-5} \cmidrule{6-7} \cmidrule{9-10} \cmidrule{11-12} \cmidrule{13-14}
 Window Size & 5 & 10 & 5 & 10 & 5 & 10 & & 5 & 10 & 5 & 10 & 5 & 10 \\\bottomrule
\\
\multicolumn{14}{c}{\it Context dictionary = the $10,000$ most frequent words} \\
\\
WS-353 & 0.48 & 0.54 & 0.54 & {\bf 0.57} & {\bf \underline{0.60}} & {\bf \underline{0.60}} & & 0.40 & 0.38 & 0.41 & 0.39 & 0.42 & 0.41 \\
RG-65 & {\bf 0.55} & 0.50 & 0.49 & {\bf \underline{0.56}} & 0.46 & {\bf 0.52} & & 0.33 & 0.32 & 0.36 & 0.33 & 0.37 & 0.36 \\
RW & 0.27 & 0.25 & {\bf 0.32} & {\bf 0.30} & {\bf \underline{0.34}} & {\bf 0.30} & & 0.21 & 0.23 & 0.21 & 0.23 & 0.21 & 0.22 \\
Syn. Ana. & 46.3 & 51.0 & 58.6 & {\bf 61.0} & {\bf \underline{61.7}} & {\bf 59.2} & & 35.0 & 39.5 & 39.8 & 44.2 & 43.4 & 47.4 \\
Sem. Ana. & 20.4 & 35.9 & 29.1 & 47.0 & 34.0 & {\bf 48.0} & & 18.1 & 23.1 & 21.6 & 29.0 & 24.8 & 32.5 \\\bottomrule
\\
\multicolumn{14}{c}{\it Context dictionary = words whose frequency is between $10^{-5}$ and $10^{-6}$} \\
\\
WS-353 & 0.46 & 0.47 & 0.54 & 0.54 & 0.54 & 0.55 & & 0.28 & 0.27 & 0.23 & 0.26 & 0.22 & 0.25 \\
RG-65 & 0.46 & 0.40 & 0.41 & 0.42 & 0.49 & 0.45 & & 0.29 & 0.31 & 0.26 & 0.30 & 0.23 & 0.29 \\
RW & 0.24 & 0.24 & 0.27 & 0.24 & 0.27 & 0.29 & & 0.19 & 0.21 & 0.15 & 0.16 & 0.11 & 0.14 \\
Syn. Ana. & 39.0 & 45.1 & 52.9 & 53.7 & 56.4 & 58.8 & & 45.2 & 47.4 & 46.1 & 48.7 & 47.3 & 49.2 \\
Sem. Ana. & 24.1 & 36.7 & 38.0 & {\bf 54.3} & 47.3 & {\bf \underline{62.5}} & & 28.8 & 37.7 & 33.9 & 42.3 & 38.9 & 46.4\\\bottomrule 
\hline
\end{tabular}
\end{center}
\caption{Performance comparison between dimensionality reduction with stochastic low-rank approximation (SLRA) and Hellinger PCA (HPCA). A symmetric context of five or ten words with scenarios (1) and (4) have been used. The best three results for each dataset are in bold, and the best is underlined. Spearman rank correlation is reported on word similarity tasks. Accuracy is reported on word analogy tasks.}
\label{sim-table}
\end{table*}

\subsubsection{Stochastic Low-Rank Approximation vs Covariance-based PCA}

In this section, we compare performance on both word evaluation tasks using the two methods for dimensionality reduction described in Section \ref{dim-reduc}.
Experiments with symmetric window of five and ten context words are run to embed word representations in a $d$-dimensional vector, with $d=\{100,200,300\}$. All results are reported in Table \ref{sim-table}.
Except for some isolated results, performance is always much better with the stochastic low-rank approximation approach than with a Hellinger PCA approach.
Calculating the reconstruction error of both approaches confirms that the PCA fails somehow to properly reduce the dimensionality. 
For a reduction from 10,000 to 100 dimensions, the PCA reconstruction error is 532.2 compared with 440.3 for the stochastic low-rank approximation.
This result is not really surprising, since it is well-known that standard PCA is exceptionally fragile, and the quality of its output can suffer dramatically in the face of only a few grossly corrupted points~\citep{Jolliffe1986}. 
Covariance-based PCA as proposed in \citet{Lebret14} is thus not an approach offering a complete guarantee of success.
An approach to robustifying PCA must be considered. This is what we propose with the stochastic low-rank approximation which, moreover, ensures a low memory consumption.
For a given dimension, a window of ten context words outperforms, in general, a window of five context words.
This confirms once again the benefit of using a larger window of context.
Performance are globally better with 300 dimensions, but performance with 200 dimensions is just slightly lower, or even better in certain cases.
Finally, using a distribution of rare words instead of frequent words (\emph{i.e.} scenario (4) instead of scenario (1) here) has only an impact on the semantic word analogy task.

\subsection{Comparison with Other Models}

\begin{table}[h]
\begin{center}
\begin{tabular}{@{}lccccc@{}}\hline\toprule
 & {\bf WS} & {\bf RG}  & {\bf RW} & {\bf SYN.} & {\bf SEM.} \\\bottomrule
\\
Raw & 0.37 & 0.31 & 0.10 & 56.8 & 83.0 \\
SLRA & 0.57 & 0.56 & 0.30 & 61.0 & 47.0 \\\bottomrule
\\
GloVe & 0.57 & 0.57 & 0.38 & 82.2 & 84.1\\
CBOW & 0.57 & 0.53 & 0.36 & 64.8 & 28.4\\
SG &  0.66 & 0.53 & 0.42 & 72.7 & 66.9\\\bottomrule
\hline
\end{tabular}
\end{center}
\caption{Comparison with raw distributions and other models for 200-dimensional word vector representations. A symmetric context window of ten words is used. Spearman rank correlation is reported on word similarity tasks. Accuracy is reported on word analogy tasks.}
\label{comp-table}
\end{table}

We compare our word representations with other available models for computing vector representations of words:
(1) the GloVe model which is also based on co-occurrence statistics of corpora~\citep{pennington2014glove}\footnote{Code available at \url{http://www-nlp.stanford.edu/software/glove.tar.gz}.},
(2) the continuous bag-of-words (CBOW) and the skip-gram (SG) architectures which learn representations from prediction-based models~\citep{Mikolov2013}\footnote{Code available at \url{http://word2vec.googlecode.com/svn/trunk/}.}.
The same corpus and dictionary $\mathcal{W}$ as the ones described in Section~\ref{wordrep} are used to train 200-dimensional word vector representations.
We use a symmetric context window of ten words, and the default values set by the authors for the other hyperparameters. 
We also compare these models directly with the raw distributions, computing similarities between them with the Hellinger distance.
Results reported in Table~\ref{comp-table} show that our approach is competitive with prediction-based models.
Using the raw probability distributions yields good results on the semantic task, while a dimension reduction with a stochastic low-rank approximation gives a better solution to compete with others on similarity and syntactic tasks.
\subsection{Inference}

\begin{table}[h]
\begin{center}
\begin{tabular}{@{}ll@{}}\hline\toprule
{\bf NEW PHRASES} & {\bf NEAREST WORDS} \\\bottomrule
 \\
 \multirow{1}{3cm}{\bf\sc british airways} & \sc\small airlines, lufthansa, qantas, \\
 & \sc\small klm, flights \\\bottomrule
 \\
 \multirow{1}{3cm}{\bf\sc chicago bulls} & \sc\small celtics, lakers, pacers, \\
 & \sc\small knicks, bulls \\\bottomrule
 \\
 \multirow{1}{3cm}{\bf\sc new york city} & \sc\small chicago, brooklyn, nyc, \\
 & \sc\small manhattan, philadelphia \\\bottomrule
 \\
 \multirow{2}{3cm}{\bf\sc president of the united states} & \sc\small president, senator, bush, \\
 & \sc\small nixon, clinton \\\bottomrule
\end{tabular}
\end{center}
\caption{Examples of phrases and five of their nearest words. Phrases representations are inferred using the encoding matrix $U$ with a symmetric window of ten context words and 300 dimensions.}
\label{neigh-table2}
\end{table}

Relying on word co-occurrence statistics to represent words in vector space provides a framework to easily generate representations for unseen words.
This is a clear advantage compared to methods focused on learning word embeddings, where the whole system needs to be trained again to learn representations for these new words.
To infer a representation for a new word $w_{\text{new}}$, one only needs to count its context words over a large corpus of text to build the distribution $\sqrt{P_{w_{\text{new}}}}$.
This nice feature can be extrapolated to phrases. 
Table \ref{neigh-table2} presents some interesting examples of unseen phrases where the meaning clearly depends on the composition of their words. For instance, words from the entity \emph{Chicago Bulls} differ in meaning when taken separately. \emph{Chicago} will be close to other american cities, and \emph{Bulls} will be close to other horned animals.
However, it can be seen in Table \ref{neigh-table2} that our model infers a representation for this new phrase which is close to other NBA teams, like the \emph{Lakers} or the \emph{Celtics}.
This also works with longer phrases, such as \emph{New York City} or \emph{President of the United States}.

\section{CONCLUSION}

We presented a systematic study of a method based on counts and the Hellinger distance for building word vector representations.
The main findings are: (1) a large window of context words is crucial to capture both syntactic and semantic information; (2) a context dictionary of rare words helps for capturing semantic, but by using just a fraction of the most frequent words already ensures a high level of performance; (3) a dimensionality reduction with a stochastic low-rank approximation approach outperforms the PCA approach.
The objective of the paper was to rehabilitate count-vector-based models, whereas nowadays all the attention is directed to context-predicting models.
We show that such a simple model can give nice results on both similarity and analogy tasks.
Better still, inference of unseen words or phrases is easily feasible when relying on counts.

\subsubsection*{Acknowledgements}

This work was supported by the HASLER foundation through the grant ``Information and Communication Technology for a Better World 2020'' (SmartWorld).

\bibliographystyle{plainnat}
\bibliography{web}

\end{document}